\pdfoutput=1

\documentclass[11pt]{article}

\usepackage{EACL2023}

\usepackage{times}
\usepackage{latexsym}

\usepackage[T1]{fontenc}

\usepackage[utf8]{inputenc}

\usepackage{microtype}

\usepackage{inconsolata}

\usepackage[english]{babel}
\usepackage[autostyle, english = american]{csquotes}
\usepackage{amsmath}
\usepackage{amsfonts}
\usepackage{textcomp}

\usepackage{enumerate}
\usepackage[inline]{enumitem}

\usepackage{graphicx}                             
\usepackage{subcaption}
\usepackage{multirow}
\usepackage{tabularx}
\usepackage{epstopdf}
\newcolumntype{Y}{>{\centering\arraybackslash}X}

\title{Divergence-Based Domain Transferability for Zero-Shot Classification}

\author{Alexander Pugantsov \and Richard McCreadie \\ School of Computing Science, University of Glasgow, UK}

\begin{document}
\maketitle
\begin{abstract}
\looseness -1 Transferring learned patterns from pretrained neural language models has been shown to significantly improve effectiveness across a variety of language-based tasks, meanwhile further tuning on intermediate tasks has been demonstrated to provide additional performance benefits, provided the intermediate task is sufficiently related to the target task. However, how to identify related tasks is an open problem, and brute-force searching effective task combinations is prohibitively expensive. Hence, the question arises, \emph{are we able to improve the effectiveness and efficiency of tasks with no training examples through selective fine-tuning?} In this paper, we explore statistical measures that approximate the divergence between domain representations as a means to estimate whether tuning using one task pair will exhibit performance benefits over tuning another. This estimation can then be used to reduce the number of task pairs that need to be tested by eliminating pairs that are unlikely to provide benefits. Through experimentation over 58 tasks and over 6,600 task pair combinations, we demonstrate that statistical measures can distinguish effective task pairs, and the resulting estimates can reduce end-to-end runtime by up to 40\%.
\end{abstract}

\section{Introduction}
\noindent \looseness -1 As the accuracy of neural models continues to increase, so does the computational cost of training and storing them. One approach of mitigating such cost is through using pretrained models to enhance performance on a downstream task, a paradigm commonly referred to as \textit{transfer learning}. However, when and why transfer learning works is not concretely understood. Traditionally, selecting the best settings, i.e. tasks and hyperparameters, for transfer often involves an extensive trial-and-error process over many combinations and can quickly make the prospect of applying transfer learning undesirable. As such, it would be valuable to estimate whether a task pair combination will be effective pre-training, i.e. estimate the \textit{transferability} of a source task to a target task.

The most optimal transferability metric would be resource-efficient, such that it is capable of accurately predicting the final performance of the model whilst minimising the amount of processing required to compute it. To this end, several works~\citep{van-asch-daelemans-2010-using, ruder-plank-2017-learning, ramesh-kashyap-etal-2021-domain} have focused on estimating transferability prior to fine-tuning, using statistical measures of divergence between the underlying feature spaces of model pairs. Domain divergence measures are used to produce a notion of distance between pairs of domains by comparing their representations and have seen significant usage in works which investigate the correlation between their estimations and performance change~\citep{van-asch-daelemans-2010-using, ramesh-kashyap-etal-2021-domain}.

Subsequent transfer learning works have also demonstrated that competitive model performance can be achieved on some target tasks even if no training samples for that task are available, an approach known as \emph{zero-data/shot learning}~\citep{larochelle2008zero}. In this work, we investigate the effectiveness of domain divergence measures in estimating the performance of zero-shot classification models, wherein models further tuned on one source task are used to directly predict on the test set of a target task without any target training samples. Specifically, we leverage the information captured by these measures as features to an auxiliary learner, whose outputs are used to rank the most effective source model for transfer to a given target task. Through the analysis of 58 sentiment classification domains, we: 
\begin{enumerate*}[label=(\arabic*)]
    \item perform a correlation analysis between each independent measure and each source-target, macro-averaged $F_{1}$-score performance output;
    \item and, for each target task, we train a series of auxiliary regression models to predict their projected performance;
    \item we then convert these into rankings of source-target pairs and evaluate the capability of our learners to find the best source model for each given target domain.
\end{enumerate*}

\section{Experiment Setup}\label{sect:experiment-setup}
\begin{figure*}[t]
  \vspace{-3mm}
  \centering
  \begin{tabular}{c @{\qquad} c }
    \includegraphics[width=.44\linewidth]{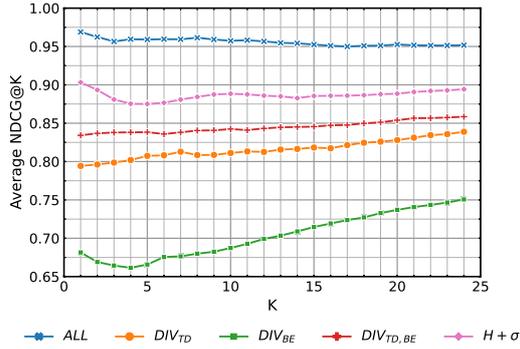} &
    \includegraphics[width=.44\linewidth]{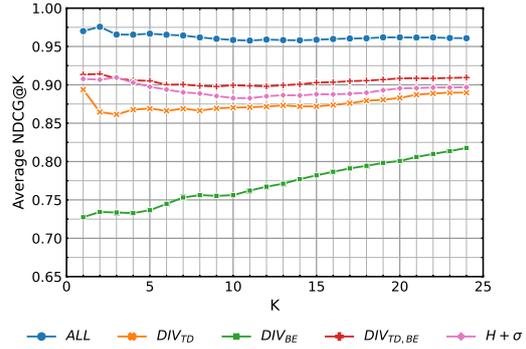} \\
    \small (a) Average NDCG@K for $N_{S}=1000$. & \small (b) Average NDCG@K for $N_{S}=25000$.
  \end{tabular}
  \caption{NDCG@K averaged across tasks for different feature sets. Higher is better. $DIV$, $H$, $\sigma$ denote divergence-, entropy-, and moments-based measures, respectively.}
  \label{fig:cw-ndcg-at-k}
\end{figure*}

\noindent \looseness -2 \noindent \looseness -2 \textbf{Measures:} \citet{ramesh-kashyap-etal-2021-domain} provide categories of divergence measures; two of which we use in our work: \textit{Geometric} measures which calculate distances between continuous representations such as word embeddings and \textit{Information-theoretic} measures which capture the distance between representations such as frequency-based distributions over co-occurring n-grams. We do not report higher-order measures as in the aforementioned work, but instead report \textit{moments}-based features, which better describe the characteristics of our individual term distributions---namely the mean, variance, skewness, and kurtosis of our distributions---as features to our learner. Following prior work \citep{tsvetkov-etal-2016-learning, ruder-plank-2017-learning}, we further complement the above measures by making use of several metrics that capture diversity and prototypicality such as entropy-based features; in our work, these measures are used with probability distributions, and are, as such, categorised here as information-theoretic. Specifically, we use the following metrics:

\begin{itemize}
    \itemsep0em 
    \item \textbf{Geometric}: Cosine distance, $l_{1}$- (or Manhattan dist.) and $l_{2}$-norm (or Euclidean dist.).
    \item \textbf{Information-theoretic}: R{\'e}nyi and Jensen-Shannon divergences~\citep{wong1985entropy, renyi1961measures}, Bhattacharyya Coeff.~\citep{bhattacharyya1943measure}, Wasserstein distance~\citep{kantorovich1960mathematical}, Entropy and R{\'e}nyi Entropy~\citep{shannon1948mathematical, renyi1961measures}, Simpson's Index~\citep{simpson1949measurement}.
    \item \textbf{Moments-based}: Mean, variance, skewness, and kurtosis ($\sigma^{n}$ where $n \in [1..4]$).
\end{itemize}

\noindent \textbf{Representations:} To compute the above metrics, we use two different representations from prior work by \citet{ruder-plank-2017-learning}, specifically 1) discrete probabilities of the most common terms across domains, using a fixed-size vocabulary $V$, where $|V| = 10,000$; and 2) a summation over probability-weighted term embeddings in each document, averaged to produce a single vector:

\begin{enumerate}[label=(\arabic*)]
    \itemsep0em 
    \item \textbf{Term Distributions (TD)}~\citep{plank-van-noord-2011-effective}: $t \in \mathbb{R}^{|V|}$ where $t_{i}$ is the probability of the $i$-th word in the vocabulary $V$.
    \item \textbf{BERT Embeddings (BE)}~\citep{Devlin2018BERT}: $\frac{1}{n} \sum_{i} v_{w_{i}} \sqrt{\frac{a}{p(w_{i})}}$ where $n$ is the number of words with embeddings in the document, $v_{w_{i}}$ is the pretrained embedding of the $i$-th term, $p(w_{i})$ its probability, and $a$ is a smoothing factor used to discount frequent probabilities. Following guidelines by \citet{ruder-plank-2017-learning}, we use this representation with geometric-based measures only, as embedding vectors can be negative. 
\end{enumerate}

Generally, since we are using these representations in a zero-shot setting, we compute divergences between the source-task training set ($D_{S}$) and the target-task test set ($D_{T}$). Entropy and moments-based measures are not used to estimate divergence between domains but used only to compute within-domain characteristics, i.e. on individual term distributions.

\vspace{2mm} \noindent \textbf{Datasets and Domains:} We make use of two ratings prediction datasets with classes in the range 1-5 and, similarly to \citet{Zhang2015CNN}, reformulate the task as a binary sentiment classification task by merging the provided labels; 1-2: negative and 3-4: positive. We focus on similar, within-task (i.e. sentiment classification) datasets to
\begin{enumerate*}[label=(\arabic*)]
\item remove task variation as a variable,
\item and to highlight the effectiveness of using statistical measures to compute divergence between similar domains which may have very minute differences in semantics and other linguistic phenomena.
\end{enumerate*} The first is the \textit{\href{https://huggingface.co/datasets/amazon_us_reviews}{Amazon Product Reviews}} dataset, using the review title and review content fields as features and divide the dataset by the product category labels. As a supplementary contribution to our work, we create the \textit{\href{https://github.com/pugantsov/multi-domain-yelp-business-reviews}{Multi-Domain Yelp Business Reviews}} dataset by extending the original \textit{Reviews} and \textit{Business} datasets provided by the \textit{\href{https://www.yelp.com/dataset}{Yelp Dataset Challenge}}, mapping top-level\footnote{We determined which categories were top-level based on \href{https://blog.yelp.com/businesses/yelp_category_list/}{an article written by Yelp}} categories of businesses to their respective reviews. After filtering out low-sample ($\leq 30,000$) domains, we have 42 and 16 domains for the Amazon and Yelp datasets, respectively.

\vspace{2mm} \noindent \textbf{Implementation Details:} We use $\text{BERT}_{base}$~\citep{Devlin2018BERT} as our base model in the experiments. With both runtime- and storage-efficiency in mind, we make use of adapter modules~\citep{pfeiffer2020AdapterHub} and train each of the domains as a source task adapter, leaving the rest of BERT's parameters frozen. More implementation and hyperparameter details can be found in Appendix~\ref{app:implementation-details}.

We divide our experiments into two separate settings by source-task sample size, $N_{S} \in [1000, 25000]$. We train 116 source-task adapters (58 $D_{S}$ $\times$ 2 $N_{S}$ settings), and evaluate a total of 6,612 source-target combinations for analysis. For our auxiliary learner, we use an XGBoost~\citep{Chen2016XGBoost} regression model. We split our training and test sets by the target task and train 2,900 regression models (for each of the 58 target domains, 2 sample sizes settings, 5 feature sets, and over 5 random seeds).

\section{Experiments and Results}\label{ref:sect-experiments}
\begin{table}[ht]
\centering
\resizebox{0.48\textwidth}{!}{
\begin{tabular}{|l|l||c|c|c|c|}
\hline
\textbf{Category}                               & \textbf{Measure}                               & \multicolumn{2}{c|}{\textbf{Term Distributions}}         & \multicolumn{2}{c|}{\textbf{BERT Embeddings}}    \\
\hline
& & 1K & 25K & 1K & 25K \\
\hline
\hline
\multirow{3}{*}{Geometric}             & Cosine Dist.                       & -0.3683* & -0.4801* & -0.3078* & -0.5792*  \\ 
\cline{2-6}
                                      & $L_{1}$ Dist.                      & -0.3699* & -0.6243* & -0.0792* & -0.4045*  \\ 
\cline{2-6}
                                      & $L_{2}$ Dist.                      & -0.3345* & -0.3551* & -0.0923* & -0.4228* \\ 
\cline{1-6}
 & R{\'e}nyi Div. & -0.4766* & -0.4273* & \multicolumn{2}{c}{}                           \\ 
\cline{2-4}
                                      & Jensen-Shannon Div.             & -0.3726* &  -0.5914* & \multicolumn{2}{c}{}                          \\
\cline{2-4}
                                      & Wasserstein Dist.      & -0.2225* & -0.3266* & \multicolumn{2}{c}{}                           \\
\cline{2-4}
Info.                                  & Bhattacharyya Coeff.             & 0.3700* & 0.5743* & \multicolumn{2}{c}{}                           \\
\cline{2-4}
Theoretic                                      & Entropy ($D_S$)                    & 0.1838* & 0.2275* & \multicolumn{2}{c}{}                           \\ 
                                      & Entropy ($D_T$)                    & -0.1603* & 0.0486* & \multicolumn{2}{c}{}                           \\ 
\cline{2-4}
                                      & R{\'e}nyi Entropy ($D_S$)   & -0.1836* &  -0.2284* & \multicolumn{2}{c}{}                          \\ 
                                      & R{\'e}nyi Entropy ($D_T$)   & 0.1618* &  -0.0503* & \multicolumn{2}{c}{}                          \\ 
\cline{2-4}
                                      & Simpson's Index ($D_S$)                     & 0.0842* & 0.1359* & \multicolumn{2}{c}{}                           \\ 
                                      & Simpson's Index ($D_T$)                     & -0.3127* &    -0.1442* & \multicolumn{2}{c}{}                        \\ 
\cline{1-4}
& $\sigma^{1}$ ($D_S$)               & -0.1321* & -0.1792* & \multicolumn{2}{c}{}                          \\ 
                                      & $\sigma^{1}$ ($D_T$)              & -0.1245* &  -0.2227* & \multicolumn{2}{c}{}                          \\ 
                                      & $\sigma^{2}$ ($D_S$)               & -0.1289* & -0.1523* & \multicolumn{2}{c}{}                           \\ 
Moments                                        & $\sigma^{2}$ ($D_T$)              & -0.1749* & -0.2549* & \multicolumn{2}{c}{}                           \\ 
Based                                      & $\sigma^{3}$ ($D_S$)               & 0.0106 & 0.0287 & \multicolumn{2}{c}{}                          \\ 
                                    & $\sigma^{3}$ ($D_T$)              & -0.3823* & -0.2643* & \multicolumn{2}{c}{}                           \\ 
                                      & $\sigma^{4}$ ($D_S$)               & 0.0006 & 0.0234 & \multicolumn{2}{c}{}                          \\ 
                                      & $\sigma^{4}$ ($D_T$)              & -0.3491* & -0.2473* & \multicolumn{2}{c}{}                          \\ 
\cline{1-4}
\end{tabular}}
\caption{\looseness -1 Spearmans $\rho$ correlations between each measure and source-target macro-averaged $F_{1}$-score performance. Asterisk denotes measure was statistically significant (\emph{P}$\leq 0.05$).}
\label{table:divergence-corr-analysis-exp1-v2}
\vspace{-3mm}
\end{table}

\begin{figure*}[t]
  \vspace{-3mm}
  \centering
  \begin{tabular}{c @{\qquad} c }
    \includegraphics[width=.45\linewidth]{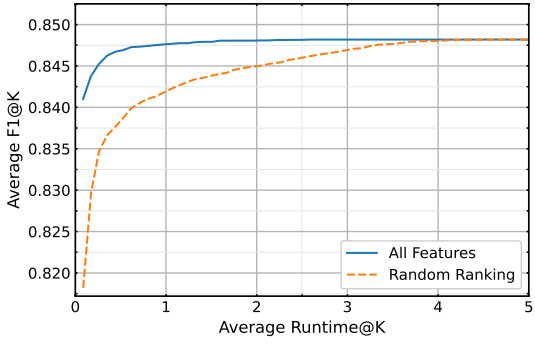} &
    \includegraphics[width=.45\linewidth]{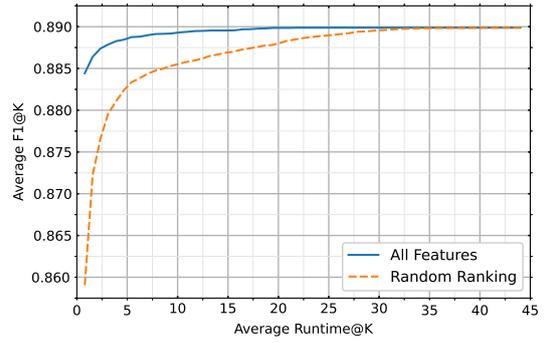} \\
    \small (a) Average F1@K vs. Runtime@K for $N_{S}=1000$. & \small (b) Average F1@K vs. Runtime@K for $N_{S}=25000$.
  \end{tabular}
  \caption{F1@K averaged across tasks vs. Total Runtime@K of source-task adapters. Higher is better. Runtime is reported in hours.}
  \label{fig:cw-f1runtime-at-k}
\end{figure*}

\noindent To evaluate whether the aforementioned statistical measures are predictive of task pair transferability, we perform a correlation analysis between the source-target pairs within each domain, where we contrast the statistical measure (which provides information about $D_S$, $D_T$, or the differences between them) and the resultant performance (measured using macro-averaged $F_{1}$) when using $D_S$ to tune a model for application on task $T$. Table~\ref{table:divergence-corr-analysis-exp1-v2} reports Spearman's Rho ($\rho$) across all sample size settings for each statistical measure. Higher correlations (distance from 0) indicate increasing predictiveness of the statistical measure of transferability. 

Using the interpretation of Spearman's Rho ($\rho$) correlation coefficients by \citet{dancey2007statistics}, we make the following observations: 
\begin{enumerate*}[label=(\arabic*)]
    \item Geometric measures exhibited a moderate-to-strong correlation for Term Distributions across both sample size settings, and strong correlations at $N_{S}=25000$ for BERT Embeddings;
    \item Between-domain Information-theoretic measures also showed moderate-to-strong performance correlations;
    \item All entropy-based measures (aside from Simpson's Index for $D_{T}$) had a weak or negligible correlation with performance;
    \item Out of all of the higher-order moments of Term Distributions, only the skewness and kurtosis of $D_{T}$ ($\sigma^{3}$ and $\sigma^{4}$) seemed to have a moderate relationship at $N_{S}=1000$, and, generally, the moments of $D_{T}$ seemed to be more correlated than that of $D_{S}$.
\end{enumerate*}

Overall, divergence measures with both representations seemed to be more predictive of source-target performances than with entropy or moments-based metrics. However, since it is unlikely that each measure was independently capable of predicting performance, we trained a series of regression models for each target task, combining these measures. Specifically, we train an XGBoost~\citep{Chen2016XGBoost} regression model (XGBRegressor) with each of the feature sets as our inputs, over five random seeds, for each of the 58 target domains and 2 sample size settings, producing 2,900 models for evaluation.

Figure~\ref{fig:cw-ndcg-at-k} shows the \textit{Average NDCG@K} values for each of these feature sets. We average the NDCG@K values across each of the 58 domains, and again over each of the 5 seeds. For both models, we achieve the best quality ranking using all of the features ($ALL$). Moreover, using divergence measures with both sets of representations ($DIV_{TD, BE}$) achieved a better ranking than using them in isolation ($DIV_{TD}$ or $DIV_{BE}$) for both settings. It is also interesting to note that the feature set containing only the entropy and moments-based ($H + \sigma$) values achieve better performance than that of those estimated via divergence measures when the source sample size is significantly limited, coinciding with patterns found in our correlation analysis (Table~\ref{table:divergence-corr-analysis-exp1-v2}); it may be the case that these features are more discriminative in cases where divergence measures are not as expressive.

\looseness -1 Finally, we evaluate the practical, downstream application of our regression models by considering how they may be used to reduce the search time in finding appropriate source models for transfer. For this experiment, we assume the user has a particular training budget $K$ to train task pairs for transfer. The more task combinations that are tried, the more likely the user is to find a better-performing model for a particular task. We use our regression models to determine the order of task pairs to be tried, using the best feature set from our prior experiments (See Fig.~\ref{fig:cw-ndcg-at-k}). We compare with a random ordering of source-task models, which we average over five random seeds to reduce variance. Figure~\ref{fig:cw-f1runtime-at-k} shows the results of our experiments. For $N_{S}=1000$, the best macro-averaged $F_{1}$ performance score over all tasks is 0.8482 which, with a grid search over all task combinations, would require 4.7 hours of training. With our approach, we can achieve a 44\% reduction in training time from 4.7 to 2.6 hours to achieve the same performance. For $N_{S}=25000$, we can achieve the maximum score of 0.8899 through a grid search of all source-target combinations at a cost of 42.4 hours of training time. With our approach, we can achieve the same score with only 24.9 hours of training or a 41\% reduction in training time. 

\looseness -1 In determining the overall runtime of our approach, we factor in the computational cost associated with generating the features required to train our regression models. Our feature generation process consists of three stages:
\begin{enumerate*}[label=(\arabic*)]
    \item the generation of term distributions and embedding representations,
    \item the computation of statistical measures in Table~\ref{table:divergence-corr-analysis-exp1-v2},
    \item and the execution of regression experiments using the $ALL$ feature set.
\end{enumerate*} A total of 232 term distributions and an equivalent number of embedding representations (58 target domains each with separate training and test sets, in two different sample size settings) were generated. The generation of both sets of representations takes 5.7 minutes at $N_{S}=1000$ and 45.9 minutes at $N_{S}=25000$. The time taken to compute all statistical measures across both representations is 3 minutes at $N_{S}=1000$ and 6.6 minutes at $N_{S}=25000$. Finally, the time taken to run the regression experiments was 5.4 minutes in total. Despite the added computational cost, our approach has resulted in a substantial reduction in end-to-end runtime, boasting a 40\% reduction at $N_{S}=1000$ and a 39\% reduction at $N_{S}=25000$, demonstrating the efficiency of our approach and the value-add of predicting which task pairs are transferable beforehand.

\section{Conclusions and Future Work}\label{ref:sect-conclusions}
\noindent In this paper, we have shown that domain divergence measures and other statistical quantities are predictive of zero-shot transferability between tasks, and that this can be used to markedly reduce time when developing effective zero-shot models. Indeed, by predicting which source-target task pairs were likely transferable pre-tuning, we were able to reduce the end-to-end time taken to find the best source-target task pairs (trained on 1,000 source-task samples) by 40\%. On the other hand, while we have demonstrated the value of using these metrics in performance estimation, there are a number of further directions worth investigating, namely: \begin{enumerate*}[label=(\arabic*)]
    \item examine the transferability across a wider range of domain and task types;
    \item investigate more complex, higher-order measures such as those outlined by \citet{ramesh-kashyap-etal-2021-domain};
    \item and to experiment with few-shot and other limited data settings.
\end{enumerate*}

\section*{Limitations}
\looseness -1 \noindent The most pronounced limitation in our work is the small variance in performance scores. As can be seen in Figure~\ref{fig:cw-f1runtime-at-k}, the difference between the lower and maximum performances is small. The difference between the minimum and maximum average performance is 0.0305 and 0.0320 for $N_{S}=1000$ and $N_{S}=25000$, respectively. Even at the individual, source-target model level, the standard deviation of performance scores at each source-task sample size setting is 0.0363 and 0.0311. As such, the benefits of zero-shot transfer are not as apparent between these domains as they would be where the domains are more textually distinct. Nevertheless, we believe it is notable that statistical measures of domain divergence and the other metrics were sufficiently capable of discerning between more effective source-task pairs, even when the domains were similar, illustrating the promise of this approach.

\bibliographystyle{acl_natbib}
\bibliography{anthology,custom}

\begin{thebibliography}{19}
\expandafter\ifx\csname natexlab\endcsname\relax\def\natexlab#1{#1}\fi

\bibitem[{Bhattacharyya(1943)}]{bhattacharyya1943measure}
Anil Bhattacharyya. 1943.
\newblock On a measure of divergence between two statistical populations
  defined by their probability distributions.
\newblock \emph{Bull. Calcutta Math. Soc.}, 35:99--109.

\bibitem[{Chen and Guestrin(2016)}]{Chen2016XGBoost}
Tianqi Chen and Carlos Guestrin. 2016.
\newblock \href {http://arxiv.org/abs/1603.02754} {Xgboost: {A} scalable tree
  boosting system}.
\newblock \emph{CoRR}, abs/1603.02754.

\bibitem[{Dancey and Reidy(2007)}]{dancey2007statistics}
Christine~P Dancey and John Reidy. 2007.
\newblock \emph{Statistics without maths for psychology}.
\newblock Pearson education.

\bibitem[{Devlin et~al.(2018)Devlin, Chang, Lee, and
  Toutanova}]{Devlin2018BERT}
Jacob Devlin, Ming{-}Wei Chang, Kenton Lee, and Kristina Toutanova. 2018.
\newblock \href {http://arxiv.org/abs/1810.04805} {{BERT:} pre-training of deep
  bidirectional transformers for language understanding}.
\newblock \emph{CoRR}, abs/1810.04805.

\bibitem[{He et~al.(2021)He, Liu, Ye, Tan, Ding, Cheng, Low, Bing, and
  Si}]{he-etal-2021-effectiveness}
Ruidan He, Linlin Liu, Hai Ye, Qingyu Tan, Bosheng Ding, Liying Cheng, Jiawei
  Low, Lidong Bing, and Luo Si. 2021.
\newblock \href {https://doi.org/10.18653/v1/2021.acl-long.172} {On the
  effectiveness of adapter-based tuning for pretrained language model
  adaptation}.
\newblock In \emph{Proceedings of the 59th Annual Meeting of the Association
  for Computational Linguistics and the 11th International Joint Conference on
  Natural Language Processing (Volume 1: Long Papers)}, pages 2208--2222,
  Online. Association for Computational Linguistics.

\bibitem[{Kantorovich(1960)}]{kantorovich1960mathematical}
Leonid~V Kantorovich. 1960.
\newblock Mathematical methods of organizing and planning production.
\newblock \emph{Management science}, 6(4):366--422.

\bibitem[{Larochelle et~al.(2008)Larochelle, Erhan, and
  Bengio}]{larochelle2008zero}
Hugo Larochelle, Dumitru Erhan, and Yoshua Bengio. 2008.
\newblock Zero-data learning of new tasks.
\newblock In \emph{AAAI}, 2, page~3.

\bibitem[{Pfeiffer et~al.(2021)Pfeiffer, Kamath, R{\"u}ckl{\'e}, Cho, and
  Gurevych}]{pfeiffer-etal-2021-adapterfusion}
Jonas Pfeiffer, Aishwarya Kamath, Andreas R{\"u}ckl{\'e}, Kyunghyun Cho, and
  Iryna Gurevych. 2021.
\newblock \href {https://doi.org/10.18653/v1/2021.eacl-main.39}
  {{A}dapter{F}usion: Non-destructive task composition for transfer learning}.
\newblock In \emph{Proceedings of the 16th Conference of the European Chapter
  of the Association for Computational Linguistics: Main Volume}, pages
  487--503, Online. Association for Computational Linguistics.

\bibitem[{Pfeiffer et~al.(2020)Pfeiffer, R\"uckl\'{e}, Poth, Kamath, Vuli\'{c},
  Ruder, Cho, and Gurevych}]{pfeiffer2020AdapterHub}
Jonas Pfeiffer, Andreas R\"uckl\'{e}, Clifton Poth, Aishwarya Kamath, Ivan
  Vuli\'{c}, Sebastian Ruder, Kyunghyun Cho, and Iryna Gurevych. 2020.
\newblock \href {https://www.aclweb.org/anthology/2020.emnlp-demos.7}
  {Adapterhub: A framework for adapting transformers}.
\newblock In \emph{Proceedings of the 2020 Conference on Empirical Methods in
  Natural Language Processing (EMNLP 2020): Systems Demonstrations}, pages
  46--54, Online. Association for Computational Linguistics.

\bibitem[{Plank and van Noord(2011)}]{plank-van-noord-2011-effective}
Barbara Plank and Gertjan van Noord. 2011.
\newblock \href {https://aclanthology.org/P11-1157} {Effective measures of
  domain similarity for parsing}.
\newblock In \emph{Proceedings of the 49th Annual Meeting of the Association
  for Computational Linguistics: Human Language Technologies}, pages
  1566--1576, Portland, Oregon, USA. Association for Computational Linguistics.

\bibitem[{Ramesh~Kashyap et~al.(2021)Ramesh~Kashyap, Hazarika, Kan, and
  Zimmermann}]{ramesh-kashyap-etal-2021-domain}
Abhinav Ramesh~Kashyap, Devamanyu Hazarika, Min-Yen Kan, and Roger Zimmermann.
  2021.
\newblock \href {https://doi.org/10.18653/v1/2021.naacl-main.147} {Domain
  divergences: A survey and empirical analysis}.
\newblock In \emph{Proceedings of the 2021 Conference of the North American
  Chapter of the Association for Computational Linguistics: Human Language
  Technologies}, pages 1830--1849, Online. Association for Computational
  Linguistics.

\bibitem[{R{\'e}nyi et~al.(1961)}]{renyi1961measures}
Alfr{\'e}d R{\'e}nyi et~al. 1961.
\newblock On measures of entropy and information.
\newblock In \emph{Proceedings of the fourth Berkeley symposium on mathematical
  statistics and probability}, 547-561. Berkeley, California, USA.

\bibitem[{Ruder and Plank(2017)}]{ruder-plank-2017-learning}
Sebastian Ruder and Barbara Plank. 2017.
\newblock \href {https://doi.org/10.18653/v1/D17-1038} {Learning to select data
  for transfer learning with {B}ayesian optimization}.
\newblock In \emph{Proceedings of the 2017 Conference on Empirical Methods in
  Natural Language Processing}, pages 372--382, Copenhagen, Denmark.
  Association for Computational Linguistics.

\bibitem[{Shannon(1948)}]{shannon1948mathematical}
Claude~Elwood Shannon. 1948.
\newblock A mathematical theory of communication.
\newblock \emph{The Bell system technical journal}, 27(3):379--423.

\bibitem[{Simpson(1949)}]{simpson1949measurement}
Edward~H Simpson. 1949.
\newblock Measurement of diversity.
\newblock \emph{nature}, 163(4148):688--688.

\bibitem[{Tsvetkov et~al.(2016)Tsvetkov, Faruqui, Ling, MacWhinney, and
  Dyer}]{tsvetkov-etal-2016-learning}
Yulia Tsvetkov, Manaal Faruqui, Wang Ling, Brian MacWhinney, and Chris Dyer.
  2016.
\newblock \href {https://doi.org/10.18653/v1/P16-1013} {Learning the curriculum
  with {B}ayesian optimization for task-specific word representation learning}.
\newblock In \emph{Proceedings of the 54th Annual Meeting of the Association
  for Computational Linguistics (Volume 1: Long Papers)}, pages 130--139,
  Berlin, Germany. Association for Computational Linguistics.

\bibitem[{Van~Asch and Daelemans(2010)}]{van-asch-daelemans-2010-using}
Vincent Van~Asch and Walter Daelemans. 2010.
\newblock \href {https://aclanthology.org/W10-2605} {Using domain similarity
  for performance estimation}.
\newblock In \emph{Proceedings of the 2010 Workshop on Domain Adaptation for
  Natural Language Processing}, pages 31--36, Uppsala, Sweden. Association for
  Computational Linguistics.

\bibitem[{Wong and You(1985)}]{wong1985entropy}
Andrew K.~C. Wong and Manlai You. 1985.
\newblock \href {https://doi.org/10.1109/TPAMI.1985.4767707} {Entropy and
  distance of random graphs with application to structural pattern
  recognition}.
\newblock \emph{IEEE Transactions on Pattern Analysis and Machine
  Intelligence}, PAMI-7(5):599--609.

\bibitem[{Zhang et~al.(2015)Zhang, Zhao, and LeCun}]{Zhang2015CNN}
Xiang Zhang, Junbo~Jake Zhao, and Yann LeCun. 2015.
\newblock \href {http://arxiv.org/abs/1509.01626} {Character-level
  convolutional networks for text classification}.
\newblock \emph{CoRR}, abs/1509.01626.

\end{thebibliography}
\newpage
\appendix

\section{Implementation Details}\label{app:implementation-details}
\noindent \textbf{Data Preparation.} For each task, we sample a holdout validation set for early stopping and a test set, both of size 2,500 (10\% of the maximum sample size), which remains fixed across both sets of experiments. After filtering by sample size, i.e. dropping domains with less than 30,000 samples, we had a total of 58 domains for comparison. Prior to sampling, the total number of domains are 43 and 22 for Amazon and Yelp datasets, respectively. We use the use the same five seeds for both data and model training.

\vspace{2mm} \noindent \textbf{Hyperparameters}. We largely follow the recommended learning rate setting of 1e-4~\citep{pfeiffer2020AdapterHub, he-etal-2021-effectiveness} for adapter training. We set the max number of epochs to 50 and an early stopping patience of 5 non-decreasing epochs. We set the maximum input length to 256 and use a batch size of 32. For the adapter configuration, we make use of the PfeifferConfig~\citep{pfeiffer-etal-2021-adapterfusion} with default settings.

\vspace{2mm} \noindent \textbf{Computing Infrastructure}. We run our experiments on 3$\times$ NVIDIA\textregistered \hspace{0.1cm} TITAN\texttrademark \hspace{0.1cm} RTX GPUs, with 130 Tensor TFLOPs of performance, 576 tensor cores, and 24 GB of GDDR6 memory. For the CPU, We use 4.5 cores of Intel\textregistered \hspace{0.1cm} Xeon\textregistered \hspace{0.1cm} Gold 5222 Processor (16.5MB Cache, 3.80 GHz) and 96GB of RAM, split across three docker containers.

\end{document}